\documentclass[10pt,twocolumn,letterpaper]{article}

\usepackage[pagenumbers]{cvpr} 

\usepackage{lineno}
\usepackage[T1]{fontenc}

\definecolor{cvprblue}{rgb}{0.21,0.49,0.74}
\usepackage[pagebackref,breaklinks,colorlinks,allcolors=cvprblue]{hyperref}

\def\paperID{5509} 
\def\confName{CVPR}
\def\confYear{2025}

\title{
    IF-MDM: Implicit Face Motion Diffusion Model \\
    for High-Fidelity Realtime Talking Head Generation
}

\author{
    Sejong Yang$^1$\thanks{This work is done as the collaboration with Yonsei University and Adobe Research.}\quad\text{ }
    Seoung Wug Oh$^2$\quad
    Yang Zhou$^2$\quad
    Seon Joo Kim$^1$ \\
    Yonsei University$^1$\quad
    Adobe Research$^2$ \\
    {\tt\small \{sejong.yang,seonjookim\}@yonsei.ac.kr   \{seoh,yazhou\}@adobe.com} \\
}

\begin{document}
\maketitle
\begin{abstract}
We introduce a novel approach for high-resolution talking head generation from a single image and audio input. Prior methods using explicit face models, like 3D morphable models (3DMM) and facial landmarks, often fall short in generating high-fidelity videos due to their lack of appearance-aware motion representation. While generative approaches such as video diffusion models achieve high video quality, their slow processing speeds limit practical application.
Our proposed model, Implicit Face Motion Diffusion Model (IF-MDM), employs implicit motion to encode human faces into appearance-aware compressed facial latents, enhancing video generation. Although implicit motion lacks the spatial disentanglement of explicit models, which complicates alignment with subtle lip movements, we introduce motion statistics to help capture fine-grained motion information. Additionally, our model provides motion controllability to optimize the trade-off between motion intensity and visual quality during inference.
IF-MDM supports real-time generation of 512x512 resolution videos at up to 45 frames per second (fps). Extensive evaluations demonstrate its superior performance over existing diffusion and explicit face models. The code will be released publicly, available alongside supplementary materials. The video results can be found on \url{https://bit.ly/ifmdm_supplementary}.
\end{abstract}
\begin{figure}[t]
    \centering
    \includegraphics[width=\linewidth]{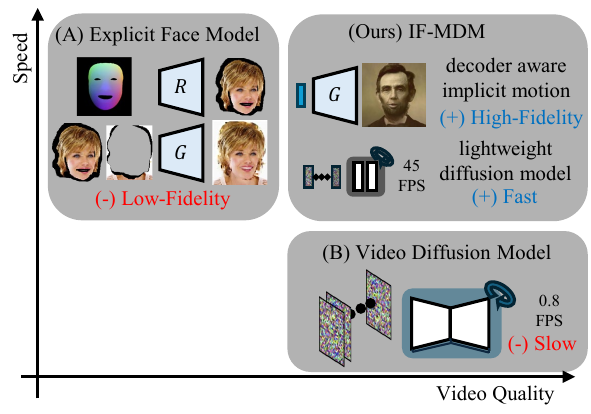}
    \caption{
        The main difference of Implicit Face Motion Diffusion Model (IF-MDM) compared to previous methods.
    }
    \label{fig:teaser}
    \vspace{-0.5cm}
\end{figure}

\section{Introduction}

Talking head generation is a task that generates talking head videos by taking a single portrait image and speech audio as input.
This field has seen significant advances driven by developments in generative models and human face models.
Generative models such as variational autoencoder (VAE)~\cite{kingma2014vae}, generative adversarial networks (GAN)~\cite{goodfellow2014gan}, and diffusion models~\cite{ho2020ddpm, rombach2022stablediffusion} have led to substantial improvements in the generation of unsupervised face video.
Building on these advancements, research has been conducted on conditional face video generation, such as lip syncing or dubbing, using models for face identity recognition and lip sync~\cite{zhong2023iplab, shen2023difftalk, liang2022gcavt, liu2022sspnerf}.

Recent advances in talking head generation have shown promising results. One line of approaches~\cite{wei2024aniportrait, stypulkowski2024diffusedheads, shen2023difftalk} train video diffusion model that directly generates talking head video conditioned on an audio signal.
While these video diffusion model approaches show impressive generation quality, they are computationally intensive making it difficult to scale to high resolutions.
Their slow processing speed also limit its pratical applicability.
Another line of approaches utilize the explicit facial model for the talking head generation conditioned on facial expression. Previous methods use facial landmarks~\cite{zhou2020makelttalk, ye2023geneface} and 3D morphable models~\cite{ye2024real3dportrait} as explicit motion representation.
While these methods are more efficient than video diffusion models, they struggle to produce high-quality video because they lack precise facial motion capture and their frame generators are unable to consistently create detailed frames with natural motion.

In this paper, we propose an implicit face motion diffusion model (IF-MDM) that address aforementioned challenges as shown in Figure~\ref{fig:teaser}.
We generally follow the motion-and-appearance disentangling approaches for efficient generation.
This includes extracting appearance information from the identity image and generating motion information from the input speech audio, then combining the appearance and motion sequence to produce a talking head video.
However, unlike previous approaches that use explicit motion representation, we learn an identity-aware implicit motion representation
Specifically, in contrast to explicit face models that rely on error-prone face warping, rendering, or transformations~\cite{xu2020deep3dportrait, ye2024real3dportrait}, our motion representation is implicit and can be directly decoded, producing high-quality talking faces with integrated appearance information.
This allows us to avoid common artifacts from previous models including the torso or background harmonization problem~\cite{ye2024real3dportrait, liu2022sspnerf}.
In addition, our motion representation is compact (a sequence of 20-dimensional vectors) allowing real-time (up to 45 FPS at 512x512 resolution) generation. 

Our system consists of two stages: 1) learning a visual encoder for disentangled motion and appearance representation and 2) generating motion sequence conditioned on driving speech audio and using it for final talking-head video rendering.
In first stage, to learn disentangled face representations, we adopt self-supervised learning framework from facial motion transfer studies~\cite{siarohin2019monkeynet, siarohin2019fomm, jeon2020crossidentity, siarohin2021mraa, wang2022lia}.
Specifically, inspired by \cite{wang2022lia}, we perform inter-frame reconstruction task that reconstruct other frames from the same video.
Only limited information for the target frame is provided through the bottleneck in the visual encoder and the linear decomposition layer, which extract compact motion details.
We use coefficients from the linear decomposition layer as the motion representation and multi-level features from the visual encoder as the appearance representation.


In the second stage, to generate natural talking head motion, we train an implicit motion generator.
First, using the encoder learned in stage one and an off-the-shelf speech encoder, we extract motion sequence and speech vector sequence from real-world talking videos.
Then, we train a diffusion transformer to generate the extracted motion sequence conditioned on the speech vector. 
During the inference, we generate audio-driven compact 20-D motion vector sequence to drive talking head animation through classifier-free guidance~\cite{ho2022classifier}.
While our model architecture is inspired by~\cite{peebles2023dit}, we made several modifications to the model to enhance the influence of the speech input. 


Despite these modifications, aligning implicit motion with speech has posed challenges, leading us to introduce a technique that leverages motion statistics.
Unlike explicit face models or video diffusion models, implicit motion is not spatially disentangled, complicating the application of commonly used methods like lip-sync loss~\cite{chung2017syncnet} in talking head generation~\cite{prajwal2020wav2lip, zhou2020makelttalk, zhou2021pcavs}.
To address this, in the second stage of training, we provided the mean and standard deviation of the motion as conditional guidance, enabling the model to use these metrics as cues.

This method also equipped our model with the capability to adjust the extent of motion during inference.
By manipulating the motion mean, we could control  head poses, rotation, and translation, whereas adjusting the motion standard deviation allowed for managing the intensity of facial expressions.

The contribution of our work is summarized as follows:

\begin{itemize}
    \item We present a framework that leverages highly compressed, appearance-aware implicit motion within a diffusion model for video generation.
    \item We introduce a methodology for controllable talking head generation using implicit motion, allowing for flexible and realistic motion representation.
    \item Finally, we provide both quantitative and qualitative comparisons with existing methods, showcasing the strengths and limitations of our approach and offering insights that may guide future research in the field.
\end{itemize}

\begin{figure*}[t]
    \centering
    \includegraphics[width=1.0\textwidth]{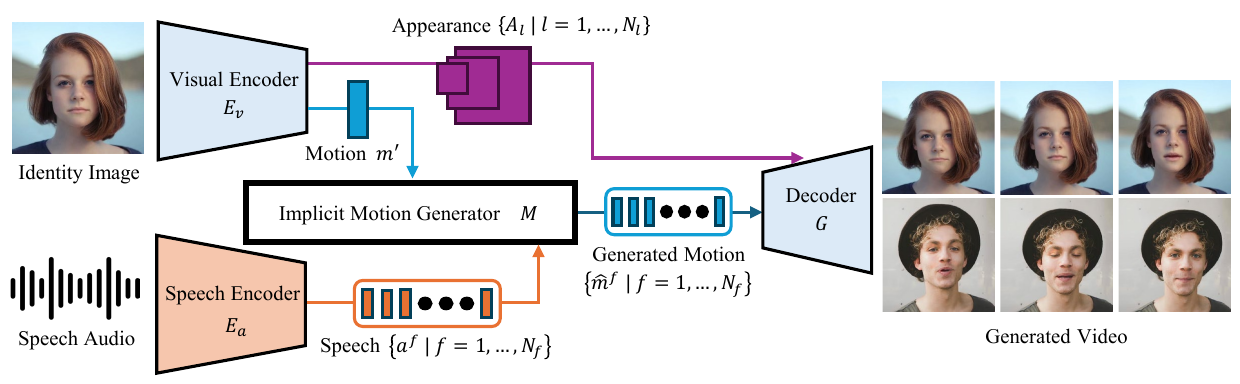}
    \caption{
        The inference pipeline of the Implicit Face Motion Diffusion Model (IF-MDM).
    }
    \label{fig:inference}
    \vspace{-0.5cm}
\end{figure*}

\section{Related Works}

\subsection{Talking head generation}

Talking head generation, which involves creating talking head videos from a single portrait image and speech audio input, has garnered significant attention in the computer vision community.
Early research focused on dubbing by utilizing encoder-decoder architectures and speech recognition models based on real talking video datasets to generate the lip region of the corresponding frame according to the given speech feature vector~\cite{prajwal2020wav2lip}.
Then, some works try to utilize the driving video and driving audio at the same time to control the head motion~\cite{zhou2021pcavs}.
Subsequently, to address the ill-posed problem of speech-to-portrait, generative models were introduced for generating the face or portrait region~\cite{zhou2020makelttalk, ye2024real3dportrait, ye2023geneface}.
Recently, methods using diffusion models have gained considerable interest~\cite{stypulkowski2024diffusedheads, shen2023difftalk, wei2024aniportrait}.
However, due to the computationally intensive nature of diffusion models, our study explores using compressed implicit motion for the diffusion model's target.

Another approach to solving talking head generation involves creating avatar models for a single identity using techniques such as neural radiance fields~\cite{guo2021adnerf, liu2022sspnerf, tang2022radnerf} or Gaussian splatting~\cite{cho2024gaussiantalker}, which generate videos based on speech input.
While this approach can produce high-quality videos, it requires retraining for each new identity, which is a notable drawback.
Our study reports that similar video quality can be achieved without such retraining.

\subsection{Motion transfer}

Motion transfer research~\cite{siarohin2019monkeynet, jeon2020crossidentity, siarohin2019fomm, siarohin2021mraa} focuses on generating a video that combines the appearance of a source image with the motion of a driving video.
In the context of the face domain, this is referred to as face reenactment~\cite{wang2022lia}.
Research in this area has involved extracting appearance information from the source image and motion information from the driving video.
Motion information has been derived using methods such as unsupervised keypoint detection, warping maps, or implicit motion vectors.
In this study, we explore the use of compressed implicit motion vectors for this purpose.

\subsection{Human avatar}

Human avatar research~\cite{su2021anerf, kwon2021nhp, saito2024relightable} focus on reconstructing and animating a realistic digital representation of a person, often using techniques like neural radiance fields (NeRF) or volumetric capture.
NeRF-based methods synthesize photorealistic images by learning the 3D structure of a scene from multiple 2D views, making them effective for capturing intricate facial details and natural expressions.
Volumetric approaches, on the other hand, utilize multiple cameras to capture the full 3D geometry of a person, resulting in high-quality avatars that can be rendered from any angle.

These methods provide high-fidelity avatar generation but often require extensive data collection and computational resources, limiting their practicality for real-time applications or for scenarios involving new identities without retraining.
Our proposed IF-MDM addresses these limitations by offering an efficient, lightweight solution that does not require retraining for each new identity, while still achieving competitive visual quality.

\begin{figure*}[t]
    \centering
    \includegraphics[width=\textwidth]{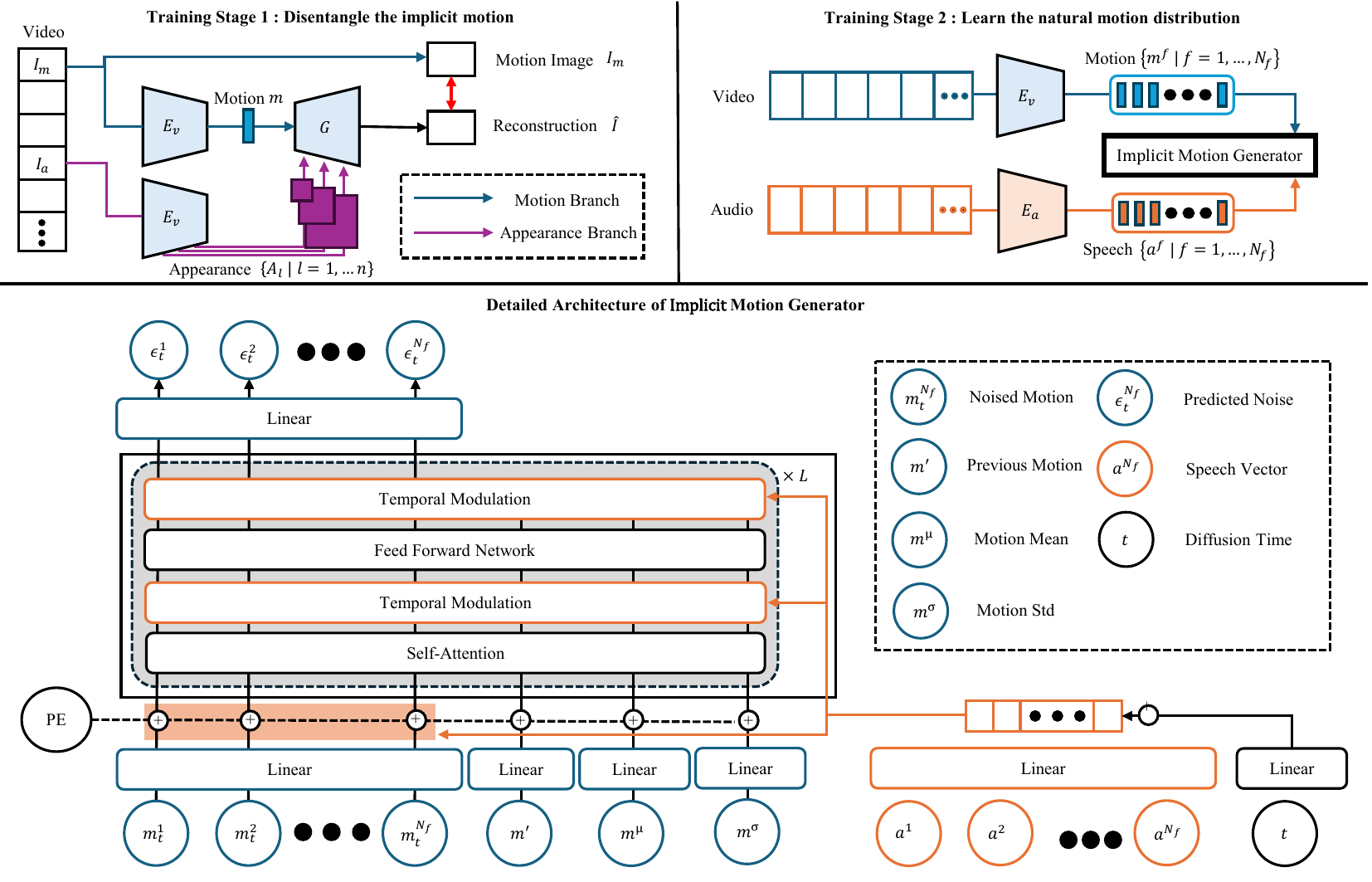}
    \caption{
        The training pipeline of our framework and the detailed architecture of the implicit motion generator.
    }
    \label{fig:training}
    \vspace{-0.5cm}
\end{figure*}
\begin{figure}[t]
    \centering
    \includegraphics[width=0.8\linewidth]{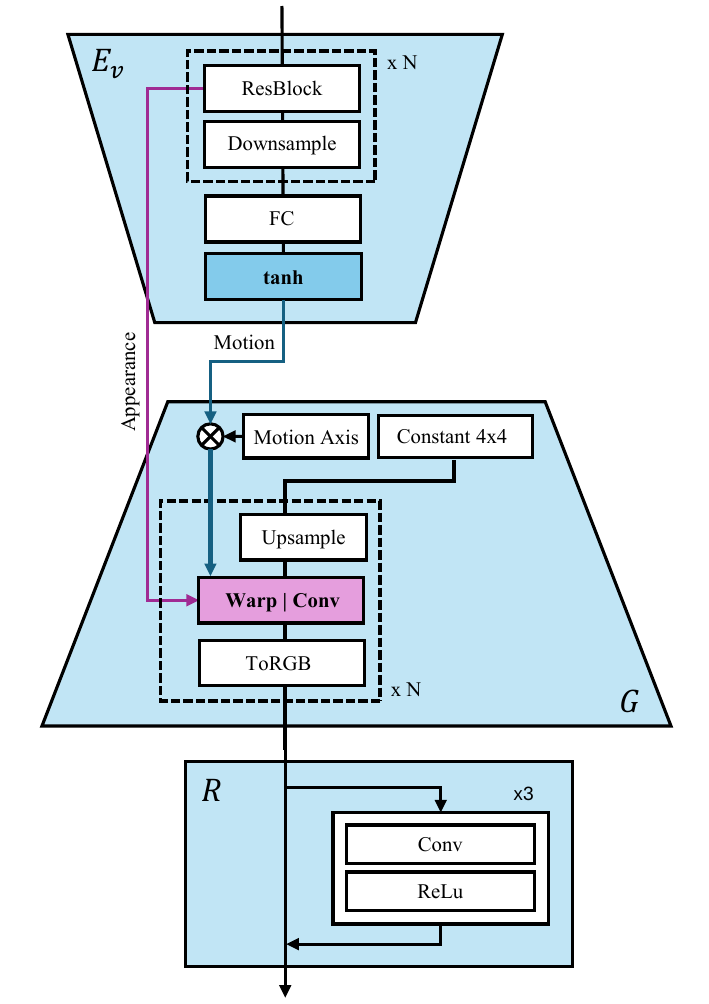}
    \caption{
        The detailed architecture of the stage 1 models.
    }
    \label{fig:stage1_model}
    \vspace{-0.5cm}
\end{figure}

\section{Method}

\subsection{Diffusion model preliminary}

In this section, we introduce the theoretical basis of the diffusion processes used in our model.
The diffusion models~\cite{ho2020ddpm} are a class of generative models that transform a data distribution into a known noise distribution through a process called the forward process, and reverse this transformation during inference to generate new data samples from noise.

The forward diffusion process is defined by a sequence of latent variables $\mathbf{x}_0, \mathbf{x}_1, \dots, \mathbf{x}_T$ where $\mathbf{x}_0$ is the data distribution and $\mathbf{x}_T$ is the noise distribution.
The transformation at each step $t$ is modeled as:

\begin{equation}
    \mathbf{x}_t = \sqrt{1 - \beta_t} \mathbf{x}_{t-1} + \sqrt{\beta_t} \mathbf{z}_t, \quad \mathbf{z}_t \sim \mathcal{N}(0, \mathbf{I})
    \label{eq:forward_diffusion}
\end{equation}

where $\beta_t$ are variance schedules that are learned during training.
The reverse process is then defined as:

\begin{equation}
    \mathbf{x}_{t-1} = \frac{1}{\sqrt{1 - \beta_t}} \left( \mathbf{x}_t - \frac{\beta_t}{\sqrt{1 - \beta_t}} \boldsymbol{\epsilon}_\theta (\mathbf{x}_t, t) \right)
    \label{eq:reverse_diffusion}
\end{equation}

where $\boldsymbol{\epsilon}_\theta (\mathbf{x}_t, t)$ is the noise predicted by the model's neural network, parameterized by $\theta$.
The goal during training is to optimize $\theta$ such that the predicted noise $\boldsymbol{\epsilon}_\theta$ closely matches the actual noise $\mathbf{z}_t$, minimizing the difference between the generated sample $\mathbf{x}_0$ and the real data distribution.

This foundational mechanism enables the generation of high-quality images by gradually refining the sample from noise over multiple timesteps, ensuring detailed and coherent outputs.
In our framework, the implicit motion generator is trained and then deployed for inference within the diffusion pipeline.

The classifier-free guidance~\cite{ho2022cfg} is a technique employed in diffusion models to enhance sample quality without the need for an additional classifier during training.
This method modifies the conditioning mechanism in the model, allowing for more flexible and controllable generation.

Mathematically, it can be described as:

\begin{equation}
    \mathbf{x}_{t-1} = \frac{1}{\sqrt{1 - \beta_t}} \left( \mathbf{x}_t - \frac{\beta_t}{\sqrt{1-\beta_t}} \boldsymbol{\epsilon}_\theta (\mathbf{x}_t, t, c + s \cdot \hat{c}) \right)
    \label{eq:classifier_free_guidance}
\end{equation}

Here, $c$ is the original condition vector, and $\hat{c}$ is an additional noise component. The scaling factor $s$ controls the intensity of guidance, enhancing the model's ability to adhere to the condition.
Typically, during inference, $s$ is set higher than during training to push the model towards generating more precise samples according to the given condition.
In our framework, all the condition like speech vector, the mean and standard deviation of motion sequence is given in the classifier-free guidance manner.

\subsection{Framework overview}

Figure~\ref{fig:inference} provides the inference pipeline with the proposed modules.
During inference, when provided with an identity image and audio speech, the visual encoder $E_v$ generates an appearance tensor list $\{ A_l \mid l = 1, \dots, n_l \}$ and an initial motion hint $m'$ where $N_l$ means the number of layers.
The speech encoder $E_a$ generated a speech vector $\{ a^f \mid f = 1, \dots, N_f \}$ where $n_f$ means the number of frames.
The implicit motion generator $M$ then synthesizes a motion sequence $\{ \hat{m}^f \mid f = 1, \dots, N_f \}$ based on the motion hint $m'$ and speech vector $\{ a^f \}$, which the generator $G$ uses alongside appearance information $\{ a^f \}$ to create a high-quality, synchronized talking head video output. 

Our IF-MDM framework is trained in two stages as shown in Figure~\ref{fig:training}.
In stage 1, the objective is to learn a disentangled representation of appearance and motion.
Given an appearance image $I_a$ and the corresponding motion image $I_m$ sampled from the same video, we extract a compressed motion vector $m$ from the motion frames.
Simultaneously, the appearance image provides skip-connection matrices that capture the appearance information $\{ a^f \}$. The visual encoder $E_v$ and the generator $G$ are trained jointly in this stage to separate appearance from motion.
Through an inter-frame reconstruction task, the model learns to generate realistic motion-transferd video by utilizing only essential appearance features from the identity image, achieving an efficient motion representation.
The stage1 model was greatly influenced by LIA~\cite{wang2022lia}, and tanh was used to normalize the motion representation for the training of the implicit motion generator.
This had the effect of normalizing the motion standard deviation. Additionally, since relying solely on warping resulted in lower generative quality, convolution layers were added to facilitate easier generation of teeth and occluded parts.
The detailed structure of the Stage 1 model is visualized in Figure~\ref{fig:stage1_model}.

The stage 2 involves training the implicit motion generator to learn the natural distribution of motion sequences driven by audio input.
Using the frozen visual encoder $E_v$ from stage 1 and the frozen speech encoder $E_a$, we extract the motion vector sequence $\{ m^f \}$ and the speech vector sequence $\{ a^f \}$ from real videos.
The implicit motion generator $M$ then trains on these sequences, with the speech vector acting as conditional guidance through classifier-free diffusion.

\subsection{Implicit motion generator}

The implicit motion generator is introduced for synthesizing expressive and synchronized talking head videos by learning a compact, decoder-aware motion sequence distribution.
Unlike explicit motion representations, such as 3DMM parameters or facial landmarks, which have predefined spatial structures, our implicit motion representation lacks spatial disentanglement, making it challenging to learn from speech-driven guidance alone, particularly for subtle lip movements.
To address this, we introduce additional conditional guidance using the motion mean ($m^\mu$) and the standard deviation ($m^\sigma$), which allows a more comprehensive understanding of the motion characteristics during training, as shown in the detailed architecture of implicit motion generator in Figure~\ref{fig:training}.

The motion mean $m^\mu$ and the motion standard deviation $m^\sigma$ serve as essential parameters to assist implicit motion generator in learning the overall dynamics of motion.
These statistics of the motion sequence $\{ m^f \}$ provide contextual information on the range and consistency of motion within each sequence of frames, allowing the model to generate coherent and realistic movements.
By incorporating these statistical parameters, implicit motion generator is able to effectively manage intensity, smoothness, and fidelity in generated motions, ensuring natural alignment between head orientation, pose stability, and expression variability.

The temporal modulation is utilized in the implicit motion generator.
Integrates the speech vector $a^{N_f}$ and diffusion time $t$ to facilitate gradual, smooth transitions in motion, ensuring that the generated video aligns with the rhythm and dynamics of the input audio.
By modulating the temporal dynamics based on these vectors, the model learns to maintain natural transitions across frames, leading to more lifelike and synchronized talking head videos.

In addition to temporal modulation, our architecture also employs a residual channel concatenation mechanism.
In this process, the speech vector $a^{N_f}$ and the noised motion vector $m_t^{N_f}$ are concatenated and passed through a linear layer which is not explicitly shown in the Figure~\ref{fig:training} to match the hidden dimensions.
The resulting vector is then added back to the original representation in residual manner, effectively enhancing the expressiveness and robustness of the motion representation.
This residual channel concatenation contributes to a more nuanced and precise control over the synthesized facial dynamics, resulting in higher-quality talking head generation.

The architecture is based on a diffusion transformer framework~\cite{peebles2023dit}, which has been modified to integrate the speech vector $a^{N_f}$, the diffusion time vector $t$, and additional statistical guidance $m^\mu$ and $m^\sigma$.
This combination allows the model to capture gradual transitions in motion that align with the input audio, leading to a lightweight yet effective system capable of producing high-fidelity outputs in real-time without compromising on expressive detail.

\subsection{Motion degree control}

In inference time, the degree of motion in the generated talking head videos can be controlled by adjusting the motion mean $m^\mu$ and the motion standard deviation $m^\sigma$, which influence different aspects of the motion dynamics.
These parameters provide flexibility in balancing expressiveness and visual fidelity based on application requirements as shown in the Table~\ref{tab:ablation_motion_degree} and Figure~\ref{fig:results_motion_degree}.

The motion mean $m^\mu$ represents the average movement characteristics in the sequence, including the orientation of the head and the overall consistency of facial expressions.
By adjusting $m^\mu$, we can control global head movements, ensuring that the generated video maintains a stable, natural posture.

A practical use of the motion mean $m^\mu$ is in maintaining or altering the consistency of poses and expressions.
If the motion mean is set to match the motion of a given input image $m'$, the model will attempt to generate motion sequence that retain a similar pose and expression, leading to a consistent output across the generated video.
This approach is particularly useful when stability in the generated video is desired.

Alternatively, by using the motion from the last frame of a previously generated motion sequence $\hat{m}^{n-1}$ as the motion mean, the model is encouraged to evolve into a new pose, allowing for dynamic changes.
This can be useful in scenarios where natural motion variation is preferred.

Moreover, it is possible to interpolate between these two strategies, using a combination of the motion mean derived from the identity image $m'$ and the last frame's motion $\hat{m}^{n-1}$.
This interpolation enables a gradual shift between consistent and evolving poses, providing greater flexibility in controlling how the subject moves throughout the video.
Such a mechanism makes our framework highly adaptable for different requirements like keeping the input image posture or generating the diverse pose and expression.

The motion standard deviation $m^\sigma$ modulates the variability of the motion, effectively controlling the intensity of facial expressions and finer details, such as the openness of the eyes and mouth.
During stage 1, we normalize the motion values using the tan hyperbolic function, which ensures that $m^\sigma$ values are standardized within a range of 0.0 to 1.0 as shown in the Figure~\ref{fig:stage1_model}. This normalization allows for consistent adjustments and predictable behavior in the expressiveness of the generated output.

Regarding the impact of modifying $m^\sigma$, smaller values of $m^\sigma$ result in less motion variability, which generally leads to higher visual quality and better lipsync accuracy.
The generated video appears more subtle, with realistic and stable expressions.
On the other hand, increasing $m^\sigma$ leads to greater motion variability, enhancing motion degree, and making the generated video more animated and expressive.
However, this comes at the expense of visual quality and lip-sync performance, as the more pronounced movements may affect the coherence of fine details.

By controlling $m^\sigma$, users can fine-tune the level of expressiveness in the generated talking head videos, making our framework adaptable to various needs, from realistic, subtle avatars to more animated and dynamic virtual characters.
\begin{table*}[t]
\centering
\caption{Quantitative results of the audio-driven talking head generation on the testset of HDTF dataset.}
\label{tab:results_hdtf}
\begin{tabular}{@{}ccccccc@{}}
\toprule
                                            & Image Quality   & Identity Preservation & Temporal Consistency    & \multicolumn{2}{c}{Lip-Sync}        & Speed          \\ 
\midrule
                                            & FID$\downarrow$ & CSIM$\uparrow$        & VideoScore-TC$\uparrow$ & LSE-D$\downarrow$ & LSE-C$\uparrow$ & FPS            \\
\midrule
Real3DPortrait\cite{ye2024real3dportrait})  & 74.68           & 0.982                 & 2.10                    & \textbf{8.23}     & \textbf{6.58}   & 10.21          \\
AniPortrait\cite{wei2024aniportrait}        & 49.13           & 0.978                 & 2.57                    & 11.56             & 3.00            & 0.88           \\
IF-MDM (ours)                               & \textbf{42.84}  & \textbf{0.984}        & \textbf{2.99}           & \textit{11.04}    & \textit{3.88}   & \textbf{30.90} \\ Ground Truth                                & 0.00            & 1.00                  & 2.71                    & 8.48              & 6.28            & -              \\ 
\bottomrule
\end{tabular}
    \vspace{-0.1cm}
\end{table*}
\begin{figure*}[t]
    \centering
    \includegraphics[width=\textwidth]{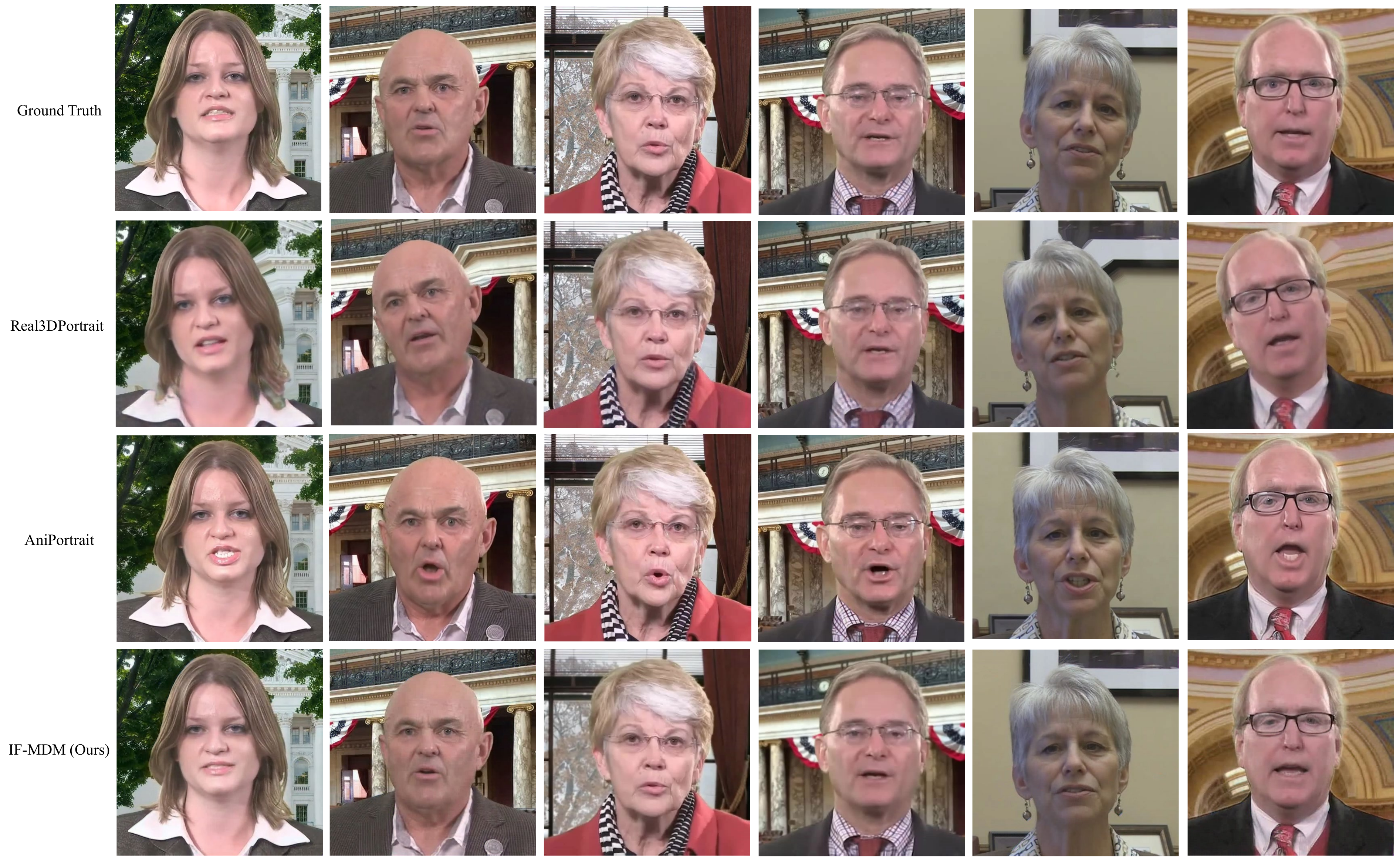}
    \caption{
        The qualitative results of HDTF datasets with baselines. The output of Real3DPortrait exhibits good lipsync quality, yet gives the impression of a floating head. The output of AniPortrait shows like heat haze in face and background when played. The video can be found on the supplementary files or \url{https://bit.ly/ifmdm_supplementary\#hdtf-title}.
    }
    \label{fig:results_hdtf}
    \vspace{-0.5cm}
\end{figure*}

\begin{table}[]
    \centering
    \caption{Ablation Study for the diffusion steps.}
    \label{tab:ablation_diffusion_steps}
    \begin{tabular}{@{}ccccc@{}}
    \toprule
    Diffusion Steps & FPS$\uparrow$   & FID$\downarrow$ & CSIM$\uparrow$ & LSE-D$\uparrow$ \\
    \midrule
    50              & 45.75           & 50.22           & 0.981          & 11.11           \\
    100             & 30.90           & 42.35           & 0.984          & 10.65           \\
    200             & 20.22           & 31.85           & 0.982          & 10.63           \\
    500             & 9.90            & 30.80           & 0.981          & 10.56           \\
    1000            & 2.18            & 31.14           & 0.980          & 10.61           \\
    \bottomrule
    \end{tabular}
    \vspace{-0.5cm}
\end{table}

\begin{figure}[t]
    \centering
    \includegraphics[width=0.5\textwidth]{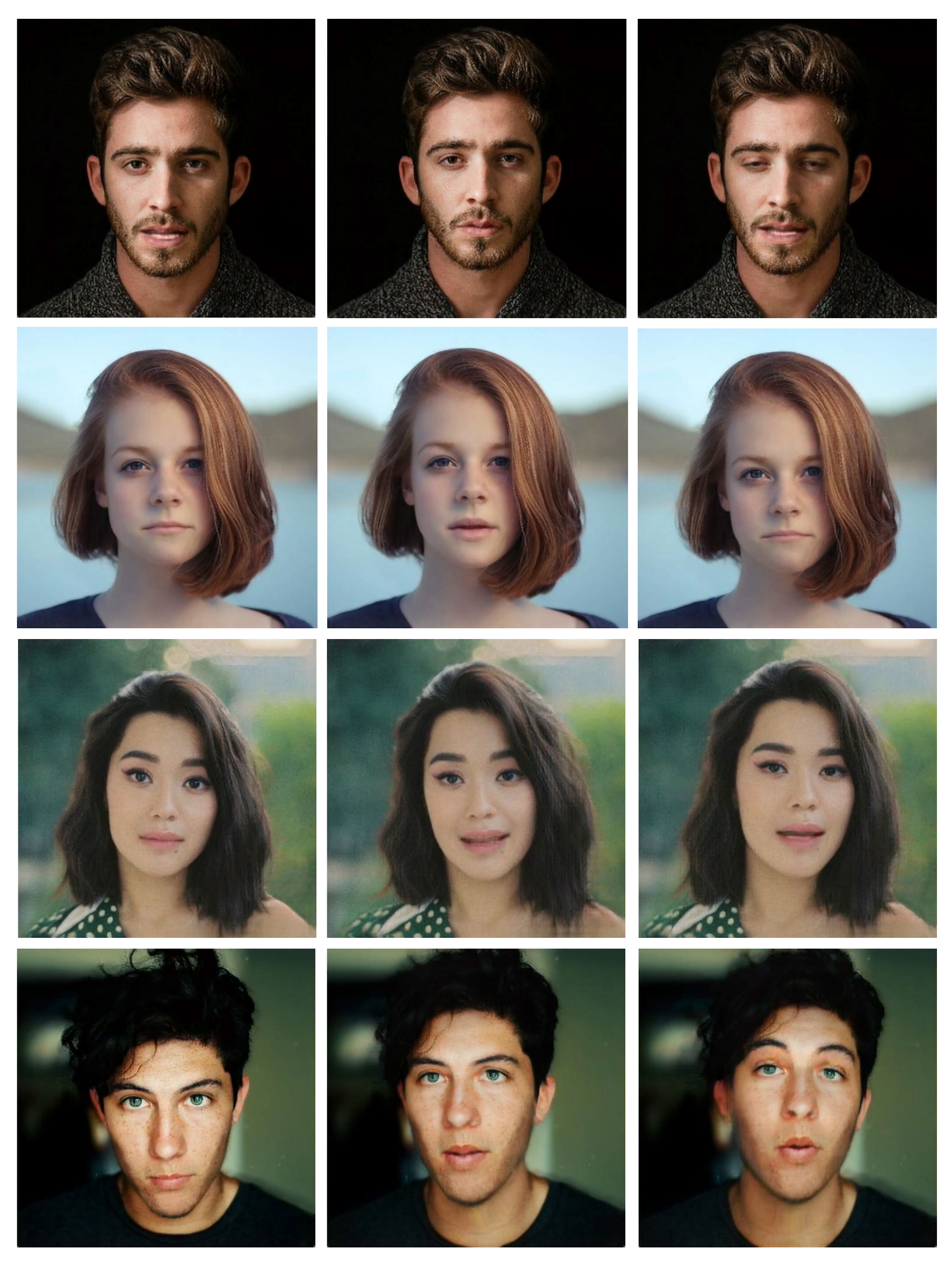}
    \caption{
        The qualitative results of the talking head generation with in the wild input image and speech audio. You can check the video and more samples in  supplementary files or \url{https://bit.ly/ifmdm_supplementary\#inthewild-title}.
    }
    \label{fig:results_inthewild}
\end{figure}

\section{Experiment}

\subsection{Implementation details}

We used the CelebV-Text~\cite{yu2022celebvtext} and HDTF~\cite{zhang2021hdtf} datasets to train our model.
For stage 1, we trained the model on video data from both datasets to learn disentangled representations of appearance and motion.
After training, the encoder was frozen and we used it to extract motion vectors.
The Wav2Vec model~\cite{baevski2020wav2vec} was used to extract speech vectors, which were then used in the implicit motion generator training.
The stage 1 training took approximately 10 days using four NVIDIA A6000 GPUs, while stage 2 training took around one week using a single NVIDIA A100 80G GPU.

\subsection{Experiment setup}

In line with previous research, we parted approximately 60 videos from the HDTF dataset as a test set for quantitative evaluation.
To evaluate the advantages of our model compared to existing methods, we selected Real3DPortrait~\cite{ye2024real3dportrait}, which uses an explicit face model, and AniPortrait~\cite{wei2024aniportrait}, a video diffusion model, for comparison.
Additionally, to assess generation performance on in-the-wild images, we downloaded and used Unsplash free license images\footnote{\url{https://unsplash.com/license}}.

\subsection{Talking head generation}

As seen in Table~\ref{tab:results_hdtf}, our method achieves over 30 fps, significantly faster than previous video diffusion model-based research, such as AniPortrait.
Additionally, it outperforms AniPortrait in both image quality and temporal consistency.
The lower temporal consistency of AniPortrait is likely due to the use of GFPGAN~\cite{wang2021gfpgan} in the official code for frame-by-frame restoration.
Furthermore, our model demonstrates superior lipsync quality than video diffusion model.

However, compared to explicit face models, our approach shows relatively lower lip sync quality.
This reflects the fact that the 3DMM utilization of Real3DPortrait results in high-quality lip sync in the rendered face, which, in turn, contributes to the high lipsync quality observed in the videos generated from this model.
However, as evident in Figure~\ref{fig:results_hdtf} and the supplementary video, the process of converting the rendered face into a portrait video gives the impression that the head is floating and unharmonized with the torso or background. 
Indeed, this results in a relatively lower FID score, highlighting the advantages and disadvantages of relying on an explicit face model.
These issues are not present in diffusion-based generative models that produce full image, such as AniPortrait and IF-MDM.
Nevertheless, our model demonstrates superior lipsync quality compared to full image generation model as shown in Table~\ref{fig:results_hdtf} and Figure~\ref{fig:results_hdtf}.

We also tested the performance of our framework on in-the-wild input images using data downloaded from the internet to obtain diverse inference results.
As shown in Figure~\ref{fig:results_inthewild}, we confirmed that our model can generate high-quality videos across various identities in-the-wild images.

\subsection{Trade-off between speed and visual quality}

To explore the trade-off between inference speed and visual quality, we conducted an ablation study on the number of diffusion steps, as presented in Table~\ref{tab:ablation_diffusion_steps}.
Our results show that increasing the number of diffusion steps generally improves visual quality, as evidenced by lower FID and higher CSIM scores.
However, this comes at the cost of reduced inference speed.

For instance, while using 100 diffusion steps provides a good balance between quality and speed with 30.90 fps, increasing to 200 steps results in higher visual fidelity but significantly lower frame rates.
Conversely, reducing the number of steps to 50 leads to faster inference at 45.75 fps but compromises visual quality.
These findings demonstrate that our model can be tuned to prioritize either speed or quality, depending on the application requirements.

\begin{table}[]
    \centering
    \caption{Ablation study for motion mean $m^{\mu}$ and motion standard $m^{\sigma}$.}
    \label{tab:ablation_motion_degree}
    \begin{tabular}{@{}cccccc@{}}
    \toprule
    $m^{\mu}$       & $m^{\sigma}$ & FID    & CSIM  & VideoScore-TC & LSE-D  \\
    \midrule
    -               & -            & 46.21  & 0.981 & 2.67          & 11.46  \\
    $m'$            & -            & 42.84  & 0.983 & 2.54          & 12.79  \\
    $\hat{m}^{n-1}$ & -            & 28.83  & 0.984 & 2.99          & 11.05  \\
    -               & 0.3          & 58.66  & 0.973 & 2.40          & 9.78   \\
    -               & 0.6          & 73.27  & 0.970 & 2.47          & 10.77  \\
    -               & 0.9          & 103.22 & 0.968 & 2.51          & 11.26  \\
    \bottomrule
    \end{tabular}
    \vspace{-0.5cm}
\end{table}
\begin{figure}[t]
    \centering
    \includegraphics[width=0.5\textwidth]{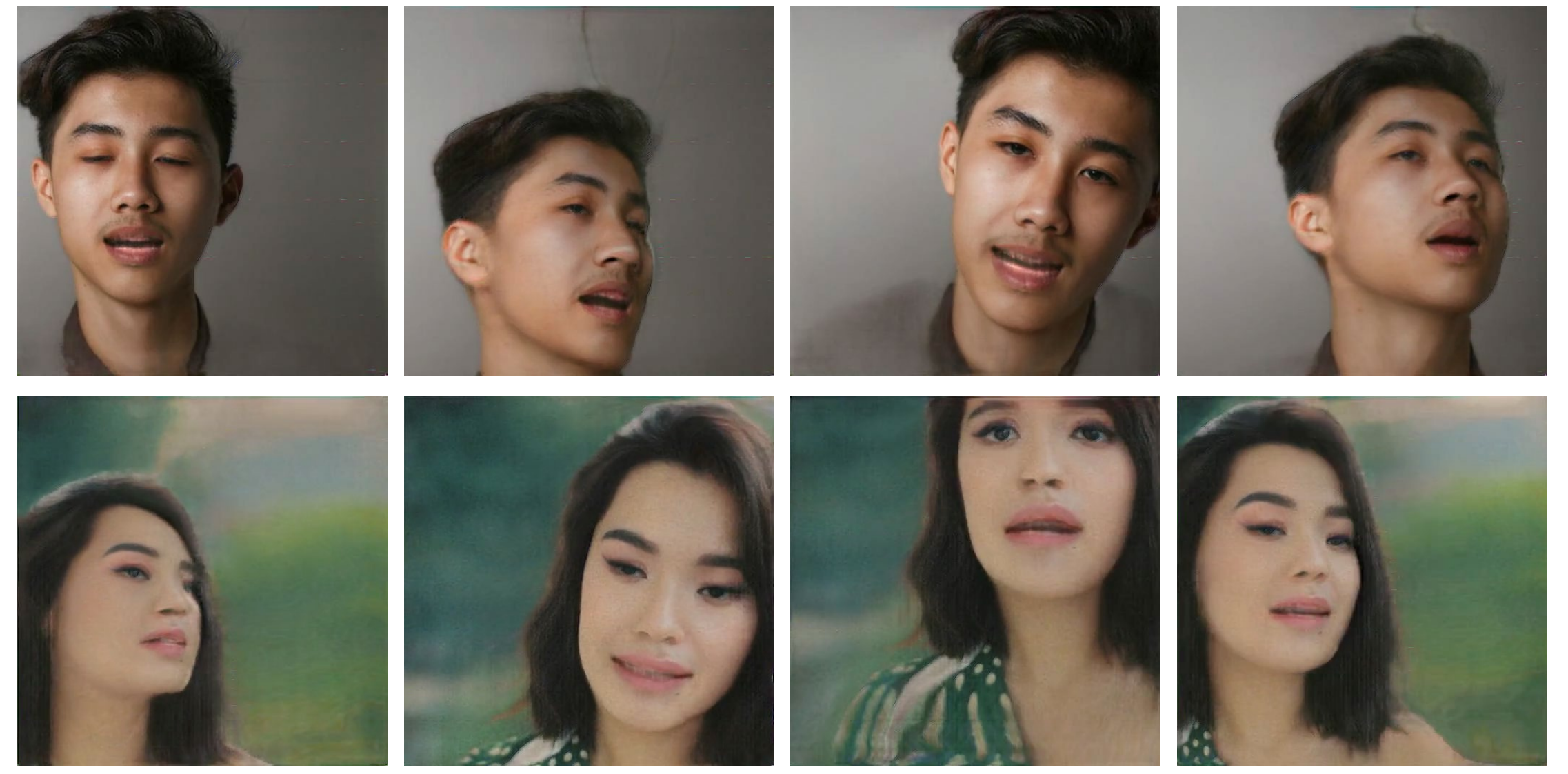}
    \caption{
        The examples with motion standard $m^\sigma$ as $0.9$. We can achieve this motion degree if we compromise the visual quality a little bit. You can check the video and more samples in supplementary files or \url{https://bit.ly/ifmdm_supplementary\#motion-degree-title}.
    }
    \label{fig:results_motion_degree}
    \vspace{-0.5cm}
\end{figure}

\subsection{Control the motion degree}

In Table~\ref{tab:ablation_motion_degree}, we conducted experiments to control the motion degree of the generated video by adjusting the motion mean $m^\mu$ and motion standard deviation $m^\sigma$.
Generally, our model performs best when provided with neutral poses, expressions, and eye and mouth openness as input images.
However, when these conditions are not met in the input image, setting the motion mean based on the motion of input image tends to produce good video quality, though with relatively lower lip sync quality, especially when using in-the-wild input images where image selection significantly impacts the results.

The motion standard $m^\sigma$ behaved as we expected.
As shown in Figure~\ref{fig:results_motion_degree}, when given a near-maximum motion standard $m^\sigma$ (0.9), the visual quality degraded considerably, and lipsync failed to work, yet the video displayed a strong motion degree with the face moving significantly across all positions in a 30-second video.
Interestingly, when the motion standard $m^\sigma$ was set to a very low value (0.3), we observed an improvement in lip sync performance.
However, other parts of the face showed minimal movement, resulting in an effect similar to dubbing~\cite{prajwal2020wav2lip}.
\section{Conclusion}

In this paper, we introduced the Implicit Face Motion Diffusion Model (IF-MDM) for high-fidelity talking head generation.
By disentangling and compressing appearance-aware face motion, IF-MDM significantly reduces computational costs while preserving video quality.
Our model achieves real-time performance, generating 512x512 resolution videos at 45 fps, which is a substantial improvement over existing methods.
Our experimental results demonstrate that IF-MDM outperforms both explicit face models and video diffusion-based approaches in terms of visual quality, identity preservation, and computational efficiency.

\section{Limitations and future works}

Future work will focus on extending the capabilities of IF-MDM to handle more complex scenarios, such as multi-person interactions and diverse environmental conditions, as well as further improving the controllability and expressiveness of the generated videos.

\section{Broader impacts}

The development of IF-MDM holds significant potential for digital media, communication, and entertainment, enabling applications in virtual assistants, digital avatars, video conferencing, and content creation.
This technology can enhance user engagement, accessibility, and personalization, making it valuable across various domains.

However, ethical concerns arise due to the potential misuse of realistic talking head generation, such as deepfakes that could lead to misinformation.
To address these risks, we advocate for the development of  responsible usage, and strong ethical guidelines.
Balancing positive applications with risk management will be key to leveraging IF-MDM responsibly.
\clearpage
{
    \small
    \bibliographystyle{ieeenat_fullname}
    \bibliography{main}

\begin{thebibliography}{35}
\providecommand{\natexlab}[1]{#1}
\providecommand{\url}[1]{\texttt{#1}}
\expandafter\ifx\csname urlstyle\endcsname\relax
  \providecommand{\doi}[1]{doi: #1}\else
  \providecommand{\doi}{doi: \begingroup \urlstyle{rm}\Url}\fi

\bibitem[Baevski et~al.(2020)Baevski, Zhou, Mohamed, and Auli]{baevski2020wav2vec}
Alexei Baevski, Yuhao Zhou, Abdelrahman Mohamed, and Michael Auli.
\newblock wav2vec 2.0: A framework for self-supervised learning of speech representations.
\newblock \emph{Advances in Neural Information Processing Systems (NeuIPS)}, 33:\penalty0 12449--12460, 2020.

\bibitem[Cho et~al.(2024)Cho, Lee, Yoon, Hong, Ko, Ahn, and Kim]{cho2024gaussiantalker}
Kyusun Cho, Joungbin Lee, Heeji Yoon, Yeobin Hong, Jaehoon Ko, Sangjun Ahn, and Seungryong Kim.
\newblock Gaussiantalker: Real-time high-fidelity talking head synthesis with audio-driven 3d gaussian splatting.
\newblock \emph{arXiv preprint arXiv:2404.16012}, 2024.

\bibitem[Chung and Zisserman(2017)]{chung2017syncnet}
Joon~Son Chung and Andrew Zisserman.
\newblock Out of time: automated lip sync in the wild.
\newblock In \emph{Asian Conference on Computer Vision Workshops(ACCV - Workshops)}, pages 251--263. Springer, 2017.

\bibitem[Goodfellow et~al.(2014)Goodfellow, Pouget-Abadie, Mirza, Xu, Warde-Farley, Ozair, Courville, and Bengio]{goodfellow2014gan}
Ian Goodfellow, Jean Pouget-Abadie, Mehdi Mirza, Bing Xu, David Warde-Farley, Sherjil Ozair, Aaron Courville, and Yoshua Bengio.
\newblock Generative adversarial nets.
\newblock \emph{Advances in Neural Information Processing Systems (NeuIPS)}, 27, 2014.

\bibitem[Guo et~al.(2021)Guo, Chen, Liang, Liu, Bao, and Zhang]{guo2021adnerf}
Yudong Guo, Keyu Chen, Sen Liang, Yong-Jin Liu, Hujun Bao, and Juyong Zhang.
\newblock Ad-nerf: Audio driven neural radiance fields for talking head synthesis.
\newblock In \emph{Proceedings of the IEEE/CVF International Conference on Computer Vision (ICCV)}, pages 5784--5794, 2021.

\bibitem[Ho and Salimans(2021)]{ho2022cfg}
Jonathan Ho and Tim Salimans.
\newblock Classifier-free diffusion guidance.
\newblock \emph{Advances in Neural Information Processing Systems Workshop (NeuIPS Workshop)}, 2021.

\bibitem[Ho and Salimans(2022)]{ho2022classifier}
Jonathan Ho and Tim Salimans.
\newblock Classifier-free diffusion guidance.
\newblock \emph{Advances in Neural Information Processing Systems Workshops (NeuIPS Workshops)}, 2022.

\bibitem[Ho et~al.(2020)Ho, Jain, and Abbeel]{ho2020ddpm}
Jonathan Ho, Ajay Jain, and Pieter Abbeel.
\newblock Denoising diffusion probabilistic models.
\newblock \emph{Advances in Neural Information Processing Systems (NeuIPS)}, 33:\penalty0 6840--6851, 2020.

\bibitem[Jeon et~al.(2020)Jeon, Nam, Oh, and Kim]{jeon2020crossidentity}
Subin Jeon, Seonghyeon Nam, Seoung~Wug Oh, and Seon~Joo Kim.
\newblock Cross-identity motion transfer for arbitrary objects through pose-attentive video reassembling.
\newblock In \emph{Proceedings of Proceedings of European Conference on Computer Vision (ECCV)}, pages 292--308. Springer, 2020.

\bibitem[Kingma(2014)]{kingma2014vae}
Diederik~P Kingma.
\newblock Auto-encoding variational bayes.
\newblock \emph{International Conference for Learning Representations (ICLR)}, 2014.

\bibitem[Kwon et~al.(2021)Kwon, Kim, Ceylan, and Fuchs]{kwon2021nhp}
Youngjoong Kwon, Dahun Kim, Duygu Ceylan, and Henry Fuchs.
\newblock Neural human performer: Learning generalizable radiance fields for human performance rendering.
\newblock \emph{Advances in Neural Information Processing Systems (NeuIPS)}, 34:\penalty0 24741--24752, 2021.

\bibitem[Liang et~al.(2022)Liang, Pan, Guo, Zhou, Hong, Han, Han, Liu, Ding, and Wang]{liang2022gcavt}
Borong Liang, Yan Pan, Zhizhi Guo, Hang Zhou, Zhibin Hong, Xiaoguang Han, Junyu Han, Jingtuo Liu, Errui Ding, and Jingdong Wang.
\newblock Expressive talking head generation with granular audio-visual control.
\newblock In \emph{Proceedings of the IEEE/CVF Conference on Computer Vision and Pattern Recognition (CVPR)}, pages 3387--3396, 2022.

\bibitem[Liu et~al.(2022)Liu, Xu, Wu, Zhou, Wu, and Zhou]{liu2022sspnerf}
Xian Liu, Yinghao Xu, Qianyi Wu, Hang Zhou, Wayne Wu, and Bolei Zhou.
\newblock Semantic-aware implicit neural audio-driven video portrait generation.
\newblock In \emph{Proceedings of Proceedings of European Conference on Computer Vision (ECCV)}, pages 106--125. Springer, 2022.

\bibitem[Peebles and Xie(2023)]{peebles2023dit}
William Peebles and Saining Xie.
\newblock Scalable diffusion models with transformers.
\newblock In \emph{Proceedings of the IEEE/CVF International Conference on Computer Vision (ICCV)}, pages 4195--4205, 2023.

\bibitem[Prajwal et~al.(2020)Prajwal, Mukhopadhyay, Namboodiri, and Jawahar]{prajwal2020wav2lip}
KR Prajwal, Rudrabha Mukhopadhyay, Vinay~P Namboodiri, and CV Jawahar.
\newblock A lip sync expert is all you need for speech to lip generation in the wild.
\newblock In \emph{ACM International Conference on Multimedia (ACMM)}, pages 484--492, 2020.

\bibitem[Rombach et~al.(2022)Rombach, Blattmann, Lorenz, Esser, and Ommer]{rombach2022stablediffusion}
Robin Rombach, Andreas Blattmann, Dominik Lorenz, Patrick Esser, and Bj{\"o}rn Ommer.
\newblock High-resolution image synthesis with latent diffusion models.
\newblock In \emph{Proceedings of the IEEE/CVF Conference on Computer Vision and Pattern Recognition (CVPR)}, pages 10684--10695, 2022.

\bibitem[Saito et~al.(2024)Saito, Schwartz, Simon, Li, and Nam]{saito2024relightable}
Shunsuke Saito, Gabriel Schwartz, Tomas Simon, Junxuan Li, and Giljoo Nam.
\newblock Relightable gaussian codec avatars.
\newblock In \emph{Proceedings of the IEEE/CVF Conference on Computer Vision and Pattern Recognition (CVPR)}, pages 130--141, 2024.

\bibitem[Shen et~al.(2023)Shen, Zhao, Meng, Li, Zhu, Zhou, and Lu]{shen2023difftalk}
Shuai Shen, Wenliang Zhao, Zibin Meng, Wanhua Li, Zheng Zhu, Jie Zhou, and Jiwen Lu.
\newblock Difftalk: Crafting diffusion models for generalized audio-driven portraits animation.
\newblock In \emph{Proceedings of the IEEE/CVF Conference on Computer Vision and Pattern Recognition (CVPR)}, 2023.

\bibitem[Siarohin et~al.(2019{\natexlab{a}})Siarohin, Lathuili{\`e}re, Tulyakov, Ricci, and Sebe]{siarohin2019fomm}
Aliaksandr Siarohin, St{\'e}phane Lathuili{\`e}re, Sergey Tulyakov, Elisa Ricci, and Nicu Sebe.
\newblock First order motion model for image animation.
\newblock \emph{Advances in Neural Information Processing Systems (NeuIPS)}, 32, 2019{\natexlab{a}}.

\bibitem[Siarohin et~al.(2019{\natexlab{b}})Siarohin, Lathuili{\`e}re, Tulyakov, Ricci, and Sebe]{siarohin2019monkeynet}
Aliaksandr Siarohin, St{\'e}phane Lathuili{\`e}re, Sergey Tulyakov, Elisa Ricci, and Nicu Sebe.
\newblock Animating arbitrary objects via deep motion transfer.
\newblock In \emph{Proceedings of the IEEE/CVF Conference on Computer Vision and Pattern Recognition (CVPR)}, pages 2377--2386, 2019{\natexlab{b}}.

\bibitem[Siarohin et~al.(2021)Siarohin, Woodford, Ren, Chai, and Tulyakov]{siarohin2021mraa}
Aliaksandr Siarohin, Oliver~J Woodford, Jian Ren, Menglei Chai, and Sergey Tulyakov.
\newblock Motion representations for articulated animation.
\newblock In \emph{Proceedings of the IEEE/CVF Conference on Computer Vision and Pattern Recognition (CVPR)}, pages 13653--13662, 2021.

\bibitem[Stypu{\l}kowski et~al.(2024)Stypu{\l}kowski, Vougioukas, He, Zi{\k{e}}ba, Petridis, and Pantic]{stypulkowski2024diffusedheads}
Micha{\l} Stypu{\l}kowski, Konstantinos Vougioukas, Sen He, Maciej Zi{\k{e}}ba, Stavros Petridis, and Maja Pantic.
\newblock Diffused heads: Diffusion models beat gans on talking-face generation.
\newblock In \emph{IEEE Winter Conference on Applications of Computer Vision (WACV)}, pages 5091--5100, 2024.

\bibitem[Su et~al.(2021)Su, Yu, Zollh{\"o}fer, and Rhodin]{su2021anerf}
Shih-Yang Su, Frank Yu, Michael Zollh{\"o}fer, and Helge Rhodin.
\newblock A-nerf: Articulated neural radiance fields for learning human shape, appearance, and pose.
\newblock \emph{Advances in Neural Information Processing Systems (NeuIPS)}, 34:\penalty0 12278--12291, 2021.

\bibitem[Tang et~al.(2022)Tang, Wang, Zhou, Chen, He, Hu, Liu, Zeng, and Wang]{tang2022radnerf}
Jiaxiang Tang, Kaisiyuan Wang, Hang Zhou, Xiaokang Chen, Dongliang He, Tianshu Hu, Jingtuo Liu, Gang Zeng, and Jingdong Wang.
\newblock Real-time neural radiance talking portrait synthesis via audio-spatial decomposition.
\newblock \emph{arXiv preprint arXiv:2211.12368}, 2022.

\bibitem[Wang et~al.(2021)Wang, Li, Zhang, and Shan]{wang2021gfpgan}
Xintao Wang, Yu Li, Honglun Zhang, and Ying Shan.
\newblock Towards real-world blind face restoration with generative facial prior.
\newblock In \emph{Proceedings of the IEEE/CVF Conference on Computer Vision and Pattern Recognition (CVPR)}, pages 9168--9178, 2021.

\bibitem[Wang et~al.(2022)Wang, Yang, Bremond, and Dantcheva]{wang2022lia}
Yaohui Wang, Di Yang, Francois Bremond, and Antitza Dantcheva.
\newblock Latent image animator: Learning to animate images via latent space navigation.
\newblock \emph{International Conference for Learning Representations (ICLR)}, 2022.

\bibitem[Wei et~al.(2024)Wei, Yang, and Wang]{wei2024aniportrait}
Huawei Wei, Zejun Yang, and Zhisheng Wang.
\newblock Aniportrait: Audio-driven synthesis of photorealistic portrait animation.
\newblock \emph{arXiv preprint arXiv:2403.17694}, 2024.

\bibitem[Xu et~al.(2020)Xu, Yang, Chen, Wen, Deng, Jia, and Tong]{xu2020deep3dportrait}
Sicheng Xu, Jiaolong Yang, Dong Chen, Fang Wen, Yu Deng, Yunde Jia, and Xin Tong.
\newblock Deep 3d portrait from a single image.
\newblock In \emph{Proceedings of the IEEE/CVF Conference on Computer Vision and Pattern Recognition (CVPR)}, pages 7710--7720, 2020.

\bibitem[Ye et~al.(2023)Ye, Jiang, Ren, Liu, He, and Zhao]{ye2023geneface}
Zhenhui Ye, Ziyue Jiang, Yi Ren, Jinglin Liu, Jinzheng He, and Zhou Zhao.
\newblock Geneface: Generalized and high-fidelity audio-driven 3d talking face synthesis.
\newblock \emph{International Conference for Learning Representations (ICLR)}, 2023.

\bibitem[Ye et~al.(2024)Ye, Zhong, Ren, Yang, Li, Huang, Jiang, He, Huang, Liu, et~al.]{ye2024real3dportrait}
Zhenhui Ye, Tianyun Zhong, Yi Ren, Jiaqi Yang, Weichuang Li, Jiawei Huang, Ziyue Jiang, Jinzheng He, Rongjie Huang, Jinglin Liu, et~al.
\newblock Real3d-portrait: One-shot realistic 3d talking portrait synthesis.
\newblock \emph{International Conference for Learning Representations (ICLR)}, 2024.

\bibitem[Yu et~al.(2023)Yu, Zhu, Jiang, Loy, Cai, and Wu]{yu2022celebvtext}
Jianhui Yu, Hao Zhu, Liming Jiang, Chen~Change Loy, Weidong Cai, and Wayne Wu.
\newblock {CelebV-Text}: A large-scale facial text-video dataset.
\newblock In \emph{Proceedings of the IEEE/CVF Conference on Computer Vision and Pattern Recognition (CVPR)}, 2023.

\bibitem[Zhang et~al.(2021)Zhang, Li, Ding, and Fan]{zhang2021hdtf}
Zhimeng Zhang, Lincheng Li, Yu Ding, and Changjie Fan.
\newblock Flow-guided one-shot talking face generation with a high-resolution audio-visual dataset.
\newblock In \emph{Proceedings of the IEEE/CVF Conference on Computer Vision and Pattern Recognition (CVPR)}, pages 3661--3670, 2021.

\bibitem[Zhong et~al.(2023)Zhong, Fang, Cai, Wei, Zhao, Lin, and Li]{zhong2023iplab}
Weizhi Zhong, Chaowei Fang, Yinqi Cai, Pengxu Wei, Gangming Zhao, Liang Lin, and Guanbin Li.
\newblock Identity-preserving talking face generation with landmark and appearance priors.
\newblock In \emph{Proceedings of the IEEE/CVF Conference on Computer Vision and Pattern Recognition}, pages 9729--9738, 2023.

\bibitem[Zhou et~al.(2021)Zhou, Sun, Wu, Loy, Wang, and Liu]{zhou2021pcavs}
Hang Zhou, Yasheng Sun, Wayne Wu, Chen~Change Loy, Xiaogang Wang, and Ziwei Liu.
\newblock Pose-controllable talking face generation by implicitly modularized audio-visual representation.
\newblock In \emph{Proceedings of the IEEE/CVF Conference on Computer Vision and Pattern Recognition (CVPR)}, pages 4176--4186, 2021.

\bibitem[Zhou et~al.(2020)Zhou, Han, Shechtman, Echevarria, Kalogerakis, and Li]{zhou2020makelttalk}
Yang Zhou, Xintong Han, Eli Shechtman, Jose Echevarria, Evangelos Kalogerakis, and Dingzeyu Li.
\newblock Makelttalk: speaker-aware talking-head animation.
\newblock \emph{ACM Transactions On Graphics (TOG)}, 39\penalty0 (6):\penalty0 1--15, 2020.

\end{thebibliography}
}

\end{document}